\pdfoutput=1

\documentclass[11pt]{article}

\usepackage[]{acl}

\usepackage{times}
\usepackage{latexsym}

\usepackage[T1]{fontenc}

\usepackage[utf8]{inputenc}

\usepackage{microtype}

\usepackage{booktabs}
\usepackage{graphicx}
\usepackage{comment}
\usepackage[]{subcaption}
\usepackage{multirow}

\newcommand{\mysubsection}[1]{\vspace{0.3em} \noindent\textbf{#1}}

\title{Using Paraphrases to Study Properties of Contextual Embeddings}

\author{Laura Burdick, Jonathan K. Kummerfeld \and Rada Mihalcea \\
  Computer Science \& Engineering \\
  University of Michigan, Ann Arbor \\
  \texttt{\{lburdick,jkummerf,mihalcea\}@umich.edu}}

\date{}

\begin{document}
\maketitle
\begin{abstract}
  We use paraphrases as a unique source of data to analyze contextualized embeddings, with a particular focus on BERT.
  Because paraphrases naturally encode consistent word and phrase semantics, they provide a unique lens for investigating properties of embeddings.
  Using the Paraphrase Database's alignments, we study words within paraphrases as well as phrase representations.
  We find that contextual embeddings effectively handle polysemous words, but give synonyms surprisingly different representations in many cases.
  We confirm previous findings that BERT is sensitive to word order, but find slightly different patterns than prior work in terms of the level of contextualization across BERT's layers.
  
\end{abstract}

\section{Introduction}

Contextualized embedding algorithms, such as BERT \cite{devlin2018bert}, have achieved impressive performance on a wide variety of tasks \cite{huang-etal-2019-glossbert,chan-fan-2019-bert,yoosuf-yang-2019-fine}.
One application of BERT is using it as a measure of sentence similarity \cite{zhang2019bertscore,sellam-etal-2020-bleurt}, based on the assumption that BERT will produce similar representations for the words in two sentences with similar semantics.

\begin{figure}
    \centering
    \includegraphics[width=\linewidth]{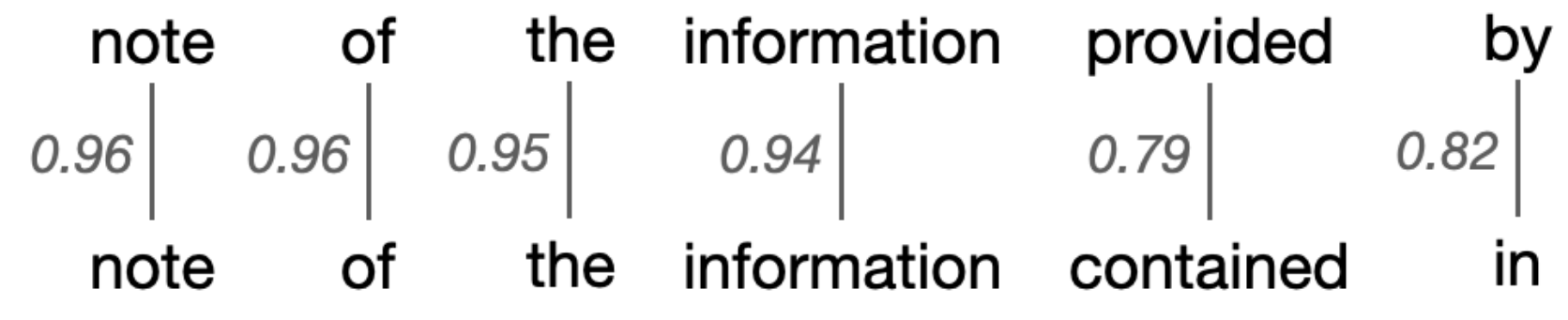}
    \caption{\label{fig:first-example}
    Example paraphrase from the PPDB with word alignment and word cosine similarities using the last layer of BERT.
    }
\end{figure}

We propose to use paraphrases with alignments between words as a tool for studying how BERT represents words and phrases.
Figure~\ref{fig:first-example} shows an example.
Critically, when considering an aligned word pair, we can assume the context has a similar impact on both words because we know the phrases are semantically similar.
Previously, paraphrases have been used to probe whether compositionality is accurately captured by BERT \cite{compositionality-test}, but we believe they can be used to explore many other questions.

Using the second version of the Paraphrase Database \cite[PPDB,][]{pavlick-etal-2015-ppdb}, we explore how consistent contextual representations are when controlling for the semantics of the context.
First, we use the human-annotated portion of the PPDB data to confirm that BERT consistently represents paraphrases.
Next, we use the highest-quality (but not all human-annotated) section of the PPDB to probe BERT's behavior in more detail.
Looking at words, BERT effectively handles variations in spelling, but does less well with spelling errors.
BERT also effectively handles words of varying levels of polysemy, but the representations for synonyms are surprisingly diverse, with a much broader distribution of similarity scores.
These findings confirm results from prior work using other methods, while uncovering new insights about contextual embedding models.

We also consider a range of other models' word representations, finding that they have similar patterns to BERT, but with aligned words that are the same receiving even more consistent representations than from BERT.
BERT gives less contextualized representations to paraphrased words than non-paraphrased words, with the exception of punctuation.
Finally, we re-evaluate work looking at patterns across BERT's layers and find that when controlling for semantics, the later layers actually produce more similar representations (in contrast to previous work).

These results show that paraphrases are a useful tool for studying representations.
By controlling for meaning while presenting interesting surface variations, they provide a unique probe of behavior.

\section{Background}

\subsection{BERTology}

There has been a growing body of research studying the inner workings of BERT and trying to quantify what it learns in various scenarios, dubbed ``BERTology''
\cite{rogers2020primer}.
Of particular interest to this paper is work that analyzes BERT's output embeddings.
Recent studies have found that embeddings created from the final layer of BERT tend to cluster according to word senses \cite{wiedemann2019does}, though this varies somewhat based on the position of a word in a sentence \cite{mickus2019you}.
The final BERT layers also produce more contextualized word embeddings than the earlier layers \cite{ethayarajh-2019-contextual}, a finding we revisit using paraphrases in Section~\ref{sec:contextualization}.

\subsection{The Paraphrase Database}

To analyze BERT, we take advantage of the unique properties of paraphrases. We use the Paraphrase Database \cite[PPDB,][]{ganitkevitch-etal-2013-ppdb,pavlick-etal-2015-ppdb}, a database of paraphrases collected using bilingual pivoting, the process of taking a particular English phrase, looking at all the foreign language phrases it can be translated into, finding all occurrences of these foreign language phrases, and then translating them back into English \cite{10.3115/1219840.1219914}.
PPDB 2.0 contains 100m+ English paraphrases, each with word alignment information, an automatically generated quality rating, and, for a subset, a human quality rating.\footnote{\href{http://paraphrase.org}{http://paraphrase.org}.} 
Word alignments are the by-product of the bilingual pivoting method used to collect the paraphrases.
When using alignments, we only consider phrases from the highest quality section of the PPDB, which are most likely to have accurate alignments. Example paraphrases with their average human annotations and automatically generated scores are shown in Table~\ref{tab:ppdb_examples}.
In general, the phrases in this dataset are short.
The longest phrases have six tokens, and the majority have fewer than six.

\begin{table}
    \centering
    \begin{tabular}{lcc}
         \toprule
         Phrases & Human & PPDB \\
         & Score & Score \\
         \midrule
         are you talking  & 1.0 & 2.7 \\
         do n't they & & \\
         \midrule
         what 's this all about ? & 4.2 & 3.9 \\
         what 's she saying ? & & \\
         \midrule
         where did they come from ? & 4.8 & 4.4 \\
         where are they from ? & & \\
         \bottomrule
    \end{tabular}
    \caption{Example tokenized paraphrases from the PPDB, with their average human annotations and automatic PPDB scores.}
    \label{tab:ppdb_examples}
\end{table}

Human quality ratings are included for 26,455 paraphrase pairs, with five annotations per paraphrase. Agreement is measured using Spearman's $\rho$ \cite{spearman1910correlation}; the average $\rho$ between two workers is 0.57, and the average $\rho$ between each worker with the other four annotators is 0.65. 

The automatic quality ratings (PPDB score) are generated by using the human annotations to fit a supervised ridge regression model. The input to the model consists of  209 hand-crafted paraphrase features, including WordNet features \cite{fellbaum1998wordnet}, distributional similarity features, and cosine similarities of generated Multiview Latent Semantic Analysis embeddings \cite{rastogi2015multiview}.
The PPDB score achieves a Spearman's $\rho$ of 0.71. In comparison, \citet{pavlick-etal-2015-ppdb} report that using the word2vec embedding of the rarest word in each paraphrase obtains Spearman's $\rho$ of 0.46.

\section{Experiments}

In our experiments, we want to use the PPDB to examine BERT's ability to consistently represent paraphrase semantics.\footnote{Note, paraphrases do not always have identical meaning. We focus on particularly similar pairs for our analysis to support our assumption that their meaning matches.}
In order to do this, we consider both phrase-level and word-level embeddings.	
Except where explicitly indicated otherwise, all experiments are run using the uncased base model of BERT, using a maximum sequence length of 128 and a batch size of 8.
We use the pretrained models provided by the Transformers library.\footnote{\url{https://huggingface.co/docs/transformers/index}}

There is a slight mismatch between the PPDB's tokenization and the format of the BERT training data. The mismatch primarily occurs with contractions and apostrophes (e.g., BERT expects ``don't'', while the PPDB is tokenized ``do n't''). This does not substantially affect the results; less than 8\% of the human-annotated paraphrase pairs contain apostrophes.
When words are broken into multiple pieces by the wordpiece tokenizer, we use the average of the pieces as the word representation.

\subsection{Phrase-Level Embeddings} \label{sec:phrase-level}

First, we consider phrase-level embeddings that capture aggregate information about all of the words in a given phrase. These embeddings show us that BERT is able to distinguish between two paraphrases, and two unrelated phrases.

We use 25,736 phrase pairs with human annotations in the PPDB.\footnote{This is 3\% smaller than the entire human-annotated subset. We were unable to map some of the human-annotated data to the data with PPDB scores (even with help from the authors of the PPDB paper).
This may be why our scores for w2v are lower than those reported by \citet{pavlick-etal-2015-ppdb}.} Each human annotation is between 1 and 5, reflecting the similarity of the two phrases.
We run each phrase through the pre-trained BERT model. For each pair of phrases, we average together the embeddings for each word to get a phrase embedding.
We create phrase embeddings using averaging because previous research has shown that this method is effective. For example, \citet{reimers-gurevych-2019-sentence} created sentence embeddings using three methods: (1) averaging word embeddings, (2) taking the maximum of word embeddings, and (3) using the CLS token vector. They found that averaging created the best sentence embeddings for semantic textual similarity tasks.

After creating phrase embeddings, we take the cosine similarity between the two embeddings.
We compare  with ground truth annotations using Spearman's $\rho$.
We do this for each of the twelve BERT layers, and the concatenation of all layers.
We use cosine similarity to compare embeddings because this metric is commonly used when working with BERT (e.g., \citet{mahmoud-torki-2020-alexu,gari-soler-apidianaki-2020-bert,kovaleva-etal-2019-revealing}).

We compare BERT to a more traditional embedding method, the continuous bag-of-words approach in word2vec (w2v) \cite{mikolov2013exploiting}. We train w2v on an English Wikipedia corpus of 5,269,686 sentences,\footnote{This data was used in \citet{tsvetkov-etal-2016-learning} and is available by contacting the authors of that paper.} using dimension size 200, a window size of five, and a minimum count of five. We choose to train w2v on Wikipedia data, in order to replicate the correlations in \citet{pavlick-etal-2015-ppdb}. We train five w2v models, using five different random seeds.\footnote{2518, 2548, 2590, 29, 401}
For each pair of phrases, we average together the embeddings for each word to get a phrase embedding, and then take the cosine similarity between the two phrase embeddings.\footnote{For both BERT and w2v, we additionally tried using the embedding of only the rarest word (with frequency measured using the full PPDB), as reported in Pavlick et al. \cite{pavlick-etal-2015-ppdb}, but this gave us consistently lower correlations.}
We report the average and standard deviation of Spearman's $\rho$ over the five models.

\begin{table*}
\small
    \centering
    \begin{tabular}{lllr}
    \toprule
    Category & \multicolumn{2}{c}{Phrases} & BERT-sim \\
\midrule
\multirow{2}{*}{High Sim} &  the transport and illegal detention of & the transportation and illegal detention of & 0.99 \\
 & representative of the secretary-general on & special representative of the secretary-general on & 0.98 \\
\midrule
\multirow{2}{*}{Low Sim} & the ohchr & the high commissioner for human rights & 0.32 \\
 & so , why & - does your shoulder bother you & 0.42 \\
\midrule
\multirow{2}{*}{Idiom} & are mad . & 're out of your mind . & 0.67 \\
& is everything all right , sir & are you okay & 0.79 \\
\bottomrule
    \end{tabular}
    \caption{\label{tab:eg-bert-sim}
    Examples of common phenomena observed in paraphrases with particularly high similarity, low similarity, and idiomatic expressions.
    }
\end{table*}

\mysubsection{Comparing Sentences and Phrases}
One difference between our work and the way BERT is normally used is that we have phrases rather than sentences.
To check that this does not substantially change BERT's behavior, we compare the embeddings for phrases in a sentence and the phrases on their own.
We take 9,780 paraphrases from the PPDB. We choose paraphrases where one of the phrases has at least six tokens, the paraphrase has a relatively good PPDB score and no syntactic placeholders. This is described further in Section~\ref{sec:word-level}. For each phrase, we find up to 100 sentences (on average, 80.5 sentences) in Gigaword \cite{graff2003english,Rush_2015}\footnote{\href{https://huggingface.co/datasets/gigaword}{https://huggingface.co/datasets/gigaword}.} and OpenSubtitles \cite{tiedemann-2012-parallel}\footnote{\href{https://opus.nlpl.eu/OpenSubtitles.php}{https://opus.nlpl.eu/OpenSubtitles.php}.} that contain that phrase.
For each sentence, we run it through BERT and average together the word embeddings for words in the phrase to create a phrase embedding.
The phrase embeddings are very similar across different sentences (average cosine similarity of $0.82\pm 0.07$).

Now we can compare (1) the average of phrase embeddings derived from sentences, with (2) embeddings for phrases in isolation, to see if BERT will be confused by not having a complete sentence.
For each phrase, we take the cosine similarity between the phrase embeddings created using these two methods.
The phrase embeddings are fairly similar (average similarity of $0.74\pm 0.12$).
This gives us confidence that BERT produces embeddings for phrases on their own that are very similar to phrases in the context of a sentence.
For the rest of our experiments, we run phrases individually through BERT, rather than in the context of complete sentences, which allows us to focus on the semantics of the phrase itself.

\begin{table}
    \small
    \centering
    \setlength{\tabcolsep}{5pt}
    \begin{tabular}{lrrrr}
        \toprule
        & \multicolumn{3}{c}{Average Length} & All \\
        Method & 1-2.5 & 2.5-4 & 4-6 & \\
        \midrule
        BERT & $0.2$ & $0.4$ & $0.51$ & $0.31$ \\
        w2v Average & $0.35$ & $0.32$ & $0.41$ & $0.43$\\
        PPDB model & $0.41$ & $0.50$ & $0.51$ &$0.50$ \\
        \midrule
        Num. phrases & $17,517$ & $5,349$ & $2,870$ & $25,736$\\
        Avg. human & $2.40$ & $2.94$ & $3.26$ & $2.60$ \\
        \bottomrule
    \end{tabular}
    \caption{\label{tab:length_breakdown} Spearman's $\rho$ between human-annotated PPDB paraphrases and different embedding methods (BERT and w2v), broken down by average paraphrase length (the average number of words in each of the two phrases in the paraphrase). Annotations are a score between 1 and 5. At the bottom of the table, we include the length distribution of the human-annotated paraphrases, as well as the average human annotation for each set of grouped lengths. For all length groupings, the w2v std. dev. is $0.0$. For paraphrases length 1-2.5, the avg. human std. dev. is $1.0$; for all other groupings, the std. dev. is $1.1$.}
\end{table}

\mysubsection{Results on the PPDB.}
Table~\ref{tab:length_breakdown} shows results for BERT, w2v, and the PPDB model, broken down by the average length of each paraphrase. For all layers, BERT improves on longer paraphrases. This is intuitive, because the longer the phrase, the more it will be able to leverage contextual information. The last layer of BERT behaves slightly differently than the other layers. While it continues to perform better on longer paraphrases, it does substantially worse on short paraphrases and slightly worse on medium paraphrases.

Similarly, w2v also improves on longer paraphrases. By taking the average of all the word embeddings for each word in the phrase, w2v has more information to incorporate into its phrase embeddings for longer paraphrases. Though w2v improves as the paraphrases grow longer, it underperforms BERT for all but the shortest paraphrases.
We also see that the automatic PPDB score does better on longer paraphrases. This could be because it incorporates distributional information, which is richer when there are more words.	
Finally, the human annotation scores show that longer paraphrases are more similar.

From Table~\ref{tab:length_breakdown}, we see that the final layer of BERT outperforms w2v and performs comparably to the PPDB score on the longest paraphrases. This is not a completely fair comparison; the PPDB model is trained specifically on this data, and has access to outside information that BERT does not, such as WordNet features and additional features derived from the translation process used to create the PPDB. These results give us confidence that BERT can distinguish between phrases that are paraphrases of each other and phrases that are not.

Looking at BERT's output, we can see several patterns in the paraphrases that receive high and low similarities.
Table~\ref{tab:eg-bert-sim} shows examples of these patterns.
For phrases with high similarity according to BERT, a single word changes or a single word is added.
On the surface, these changes have very little impact on the meaning, though the addition of the word `special' in the second case could change who is being referred to.
For phrases with low similarity according to BERT, they frequently required world knowledge (e.g., definition of an acronym) or appeared to be errors.
We also observed idioms getting reasonably high scores, but not as high as the literal paraphrases.

\emph{Conclusion: The standard way of using BERT to produce a representation of a phrase is consistent with human scores of paraphrases. All layers are effective, though the last layer struggles with shorter phrases.}

\subsubsection{One-Word Paraphrases}

In Section~\ref{sec:phrase-level}, we saw that BERT does not do as well on short phrases as it does on longer ones.
We explore the extreme case of single word paraphrases here.
Among the subset of one-word paraphrases, there is a wide range of human annotations (average annotation $2.27\pm 0.99$). To explore this further, we focus on one-word paraphrases with a human annotation of 5, the highest annotation score, indicating that these are the strongest synonyms. Among these high-quality synonyms, cosine similarities are consistently high for the last layer of BERT (average similarity $0.76\pm 0.12$).

\begin{table}
    \centering
    \begin{tabular}{llc}
        \toprule
        Phrase 1 & Phrase 2 & Cos. Sim. \\
        \midrule
        laboratoires & laboratories & 0.51 \\
        completly & totally & 0.51 \\
        fervor & enthusiasm & 0.52 \\
        79.0 & seventy-nine & 0.53 \\
        approximatly & around & 0.54 \\
        \midrule
        -mom & -mother & 0.91 \\
        1.350 & 1.35 & 0.92 \\
        characterises & characterizes & 0.92 \\
        km & kilometres & 0.92 \\
        garbage & trash & 0.96 \\
        \bottomrule
    \end{tabular}
    \caption{Cosine similarity scores for the last layer of BERT for one-word paraphrases with the highest human annotation score.}
    \label{tab:one_word_ppdb}
\end{table}

Table~\ref{tab:one_word_ppdb} shows synonyms with both the highest and lowest BERT similarities.
Misspelled words (e.g., \texttt{completly}, \texttt{approximatly}) and pairs that involve different languages (e.g., French \texttt{laboratoires}) have low cosine similarities.
Numbers appear on both the low end (e.g., \texttt{79.0} and \texttt{seventy-nine}) and the high end (e.g., \texttt{1.350} and \texttt{1.35}) of the similarity spectrum.
One difference between the similar and dissimilar number pairs is that in the similar case they both use digits, while in the dissimilar case, one uses digits while the other uses words.

\emph{Conclusion: Looking at single words shows that BERT struggles to identify synonyms, and does particularly poorly with misspellings and cross-lingual comparisons.}

\subsection{Word-Level Embeddings} \label{sec:word-level}

PPDB provides alignments between words in the paraphrases, automatically generated as part of bilingual pivoting.
We use these alignments to consider four different sets of words:

 \begin{figure}
     \centering
     \includegraphics[width=\linewidth]{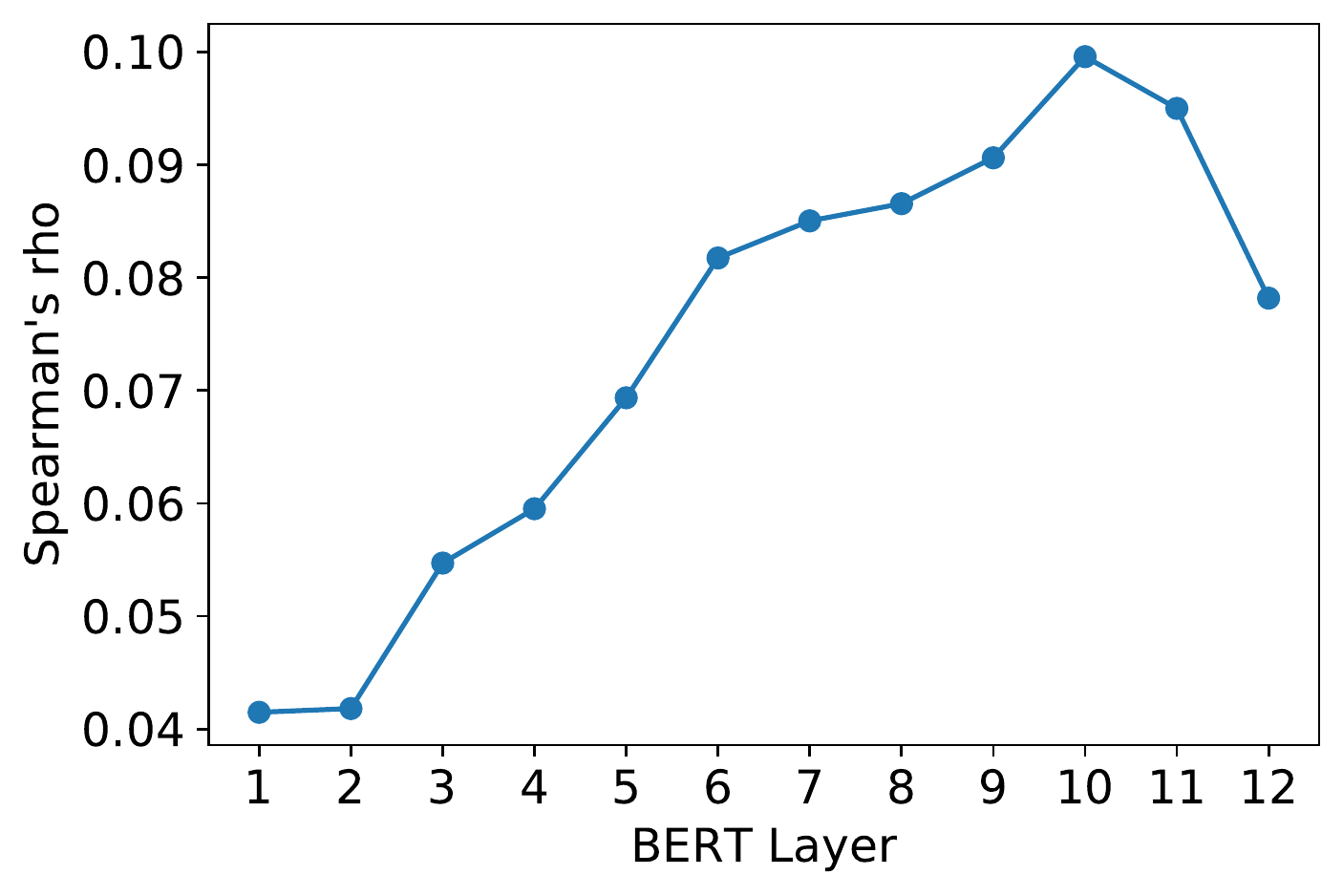}
     \caption{Spearman's $\rho$ between BERT cosine similarities and PPDB scores for all aligned same words, broken down by BERT layer.}
     \label{fig:wordlevel_correlation}
 \end{figure}

\begin{description}
\itemsep-3pt
    \item[Same, Aligned] Words that are the same in both phrases and aligned.
    \item[Same, Unaligned] Words that are the same in both phrases, but not aligned.
    These tend to be function words.
    90\% of our examples are one of (the, of, ", to, i, in, that, as, what).
    This category may have more examples of other word types if longer paraphrases are considered in future work.
    \item[Different, Aligned] Words that are aligned, but not the same.
    This case covers synonyms.
    \item[Different, Unaligned] Words that are not aligned and not the same (but still one from each phrase in a paraphrase pair).
    Note, these words are not completely unrelated.
    They are from the same paraphrase, making them more related than words from two random phrases.
\end{description}

\begin{figure}
    \centering
    \includegraphics[width=\linewidth]{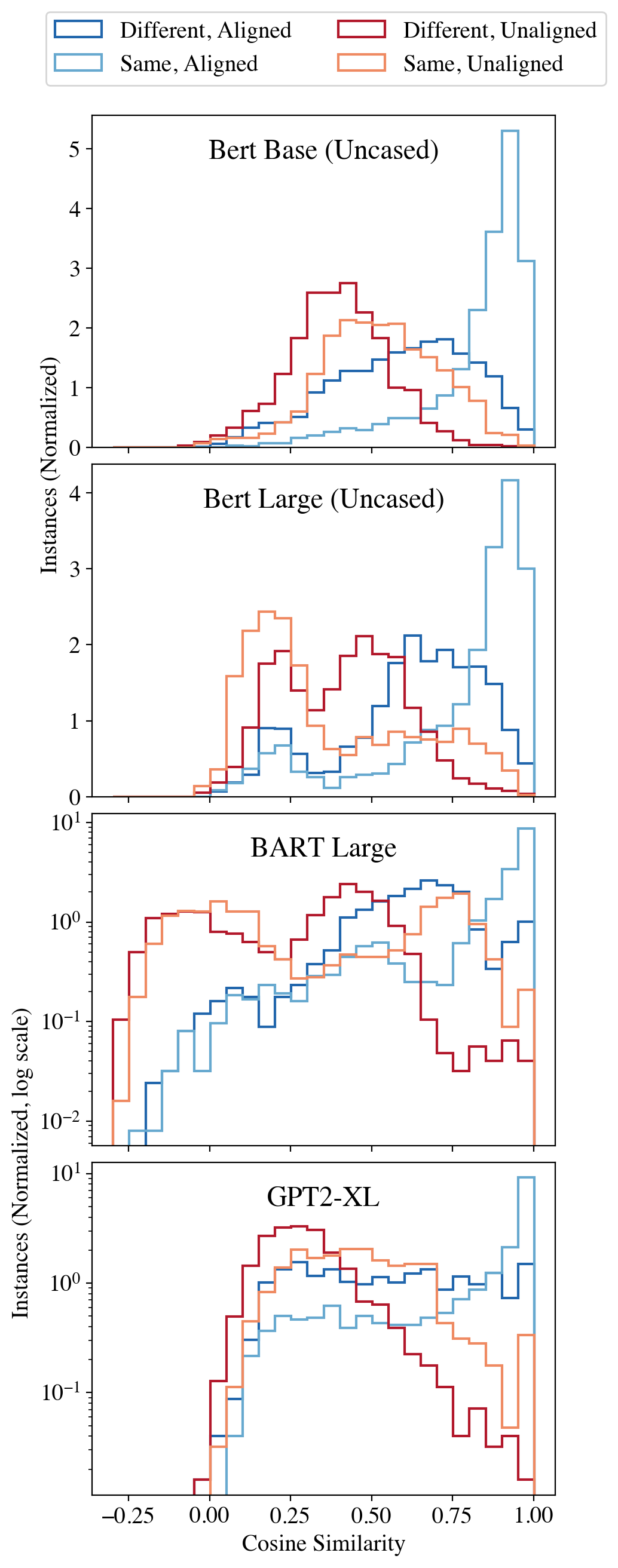}
    \caption{Distributions of cosine similarities for several models for same and different aligned and unaligned words (best seen in color). Cosine similarity is binned into bins of width 0.05. Note that the bottom two graphs use a log scale because the peak at 1.0 makes it hard to see variations otherwise.}
    \label{fig:wordComparison}
\end{figure}

In this section, we go beyond the human-annotated data considered in the previous section.
We restrict our experiments to the highest quality paraphrases in the PPDB: dataset S.
We also only consider long paraphrases (where one of the phrases has at least six tokens), and paraphrases that have no syntactic placeholders (a subset of the PPDB contains general syntactic symbols, such as \texttt{wishes to be [VP/NP]}). From this set, we randomly sample 4,000 paraphrases.  
Our sample yields 22,751 aligned same words, 25,973 aligned different words, 2,782 unaligned same words, and 163,474 unaligned different words.
We randomly sample 2,500 words from each category.
For the aligned words, we only use cases where there is a 1-1 alignment.

To generate word-level embeddings, we run each phrase through a set of transformer models and for each pair of words, we take the cosine similarity between the embeddings of the two words.

\subsubsection{Results}

Figure~\ref{fig:wordComparison} shows the distributions of similarity scores for all four sets of words for several models.
Same, aligned words consistently have the highest similarity.
The other categories tend to overlap.
Because we are using paraphrases, we would hope that aligned different words would have higher similarity, but that is not consistently the case.

Comparing the models, there are some notable variations.
Between BERT base and BERT large, the biggest shift is that unaligned words that are the same have much lower similarity in BERT large, though there is also a new peak for aligned words around 0.2.
Comparing the two BERT models with BART and GPT-2, there is a much sharper peak for BART and GPT-2 for same aligned words, which is consistent with prior work \cite{ethayarajh-2019-contextual}.

BART is the only model to have a substantial number of negatively correlated word pairs.
Many of these involve function words or punctuation.
For the unaligned cases, negative cosine similarity is fine because the words should not have the same meaning.
For the 26 cases of aligned pairs, it is unclear why the representations are so different.
For example, \texttt{the} plays the same role in \texttt{( , the commission considered} and \texttt{, the commission had before it}.
Similarly, \texttt{aim} and \texttt{view} should be very similar in \texttt{aim of improving the} and \texttt{with a view to improving the}.

Qualitatively looking at examples, we notice that when a token appears in a different position in the paraphrase, the similarity tends to be on the lower end of the distribution (e.g., \texttt{action} in the phrases \texttt{plans of action for the implementation} and \texttt{action plan for the implementation} has a similarity of 0.28).
To explore this, we consider 2,181 aligned same words.
We measured the cosine similarity of the last layer of BERT broken down by the variation in position (plotted in 
Figure~\ref{fig:words_apart} in the Appendices).
Spearman's $\rho=-0.29$ ($p$-value $<10e-42$), indicating that similarity decreases for larger changes in position.
This supports observations in prior work \citep{mickus2019you}, but now with the knowledge that the overall context has the same meaning.
This is not intuitive behavior; because these words are aligned in a paraphrase, we would expect that the position of the word would not substantially affect its representation.
This may indicate that the representations are encoding some information about syntactic structure, which can vary without changing semantics.

\emph{Conclusions: (1) Contextual word embedding methods consistently handle aligned words in paraphrases, but with substantial variations across models in how peaked the distributions of same-aligned words are. (2) Even when controlling for the meaning of the context, BERT represents words differently depending on their position.}

\begin{figure}
    \centering
\includegraphics[width=0.95\linewidth]{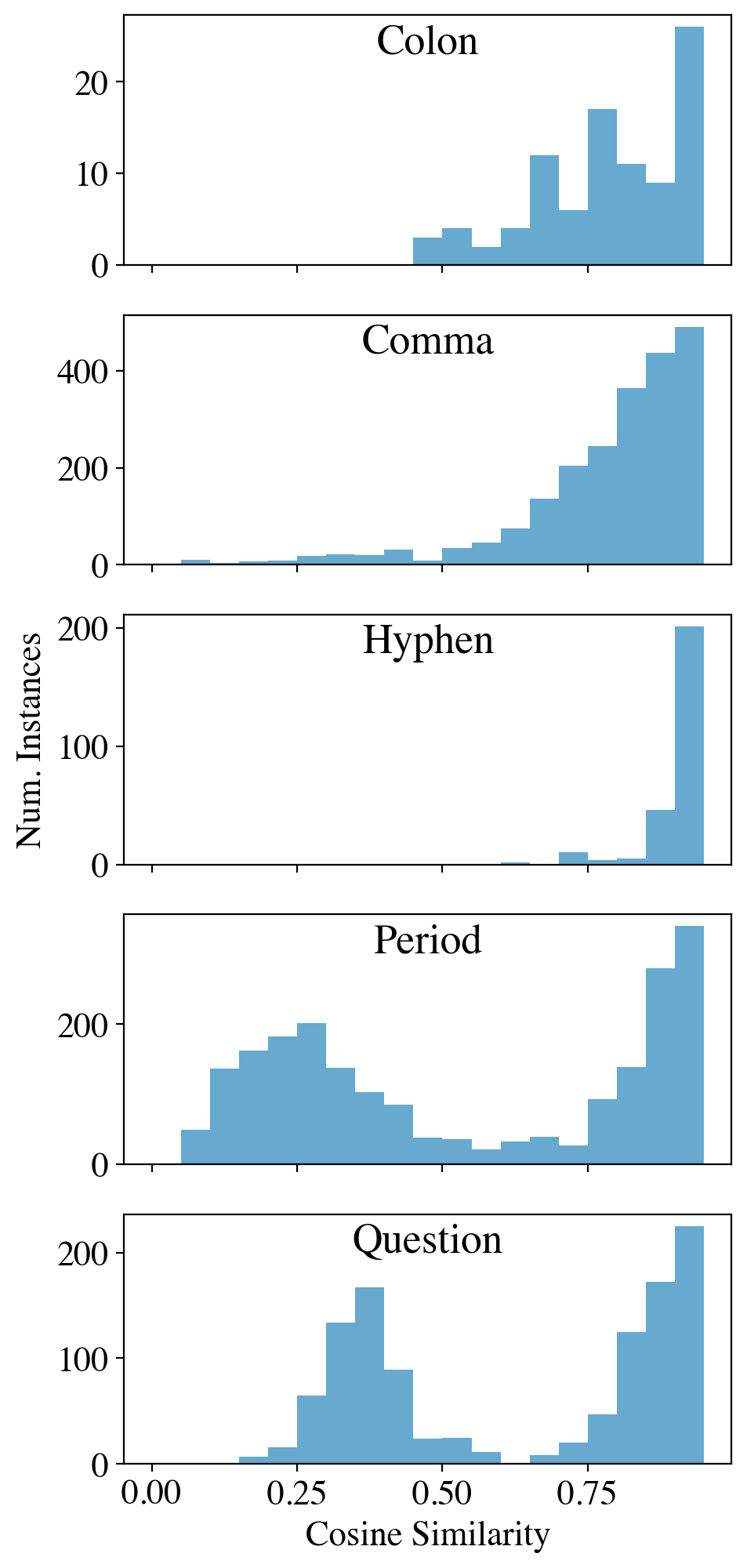}

    \caption{Distribution of cosine similarities using the last layer of BERT for aligned same words, broken down by punctuation mark.}
    \label{fig:punctuationMarks}
\end{figure}

\subsubsection{Punctuation}\label{sec:punctuation}

Punctuation is a core part of language that functions quite unlike words; punctuation groups words together or separates them, and contributes to the overall structure and meaning of a phrase or sentence. 
Punctuation plays an important role in distinguishing between different types of text, such as texts by different authors \cite{soler-company-wanner-2017-relevance} or texts produced by different Twitter communities \cite{tatman-punctuation}. Embeddings are used to generate punctuation for text that is lacking punctuation, such as recorded transcripts \cite{8682260}. 
To explore how BERT handles punctuation, we consider the cosine similarity distribution for different sets of punctuation tokens for the last layer of BERT.
We find that punctuation has a broader distribution of cosine similarities than other tokens, indicating that punctuation embeddings vary widely dependent on the surrounding context.

In Figure~\ref{fig:punctuationMarks}, we break these trends down by individual punctuation marks, focusing only on aligned same words. We look at the most common punctuation marks. Of these punctuation marks, the comma and period show the widest distributions. Even when they play the same role in the paraphrase, they can be given very different embeddings, indicating how highly contextualized these punctuation marks are. The question mark and dash are less contextualized; this is most likely because these punctuation marks are used in more prescribed circumstances.
In this dataset, in all but one example, the question mark is the last token; the dash is the first token in all but two examples.

Table~\ref{tab:punc_examples} shows examples in context of each of these punctuation marks.
Looking at the low similarity cases, one common pattern is that the phrase contains a contraction that is expanded in one phrase (e.g., ``it is'' and ``it's'').

\emph{Conclusion: BERT's representation of punctuation is surprisingly context sensitive, with substantial variation even when we control for meaning.}

\begin{table}
    \centering
    \begin{tabular}{lr}
        \toprule
        Phrases & Cos. Sim. \\
        & (Last Layer) \\
        \midrule
        it is important , however \underline{,} & 0.08 \\
        however \underline{,} it should be \\
        \midrule
        okay \underline{,} i 'm sorry & 0.84 \\
        oh \underline{,} i am so sorry \\
        \midrule
        well , it 's true \underline{.} & 0.19 \\
        this is true \underline{.} & \\
        \midrule
        , that 's all right \underline{.} & 0.15 \\
        , this is good \underline{.} & \\
        \midrule
        where have you come from \underline{?} & 0.94 \\
        where are you from \underline{?} & \\
        \midrule
        news - politics \underline{-} world - & 0.71 \\
        news - international politics \underline{-} & \\
        \bottomrule
    \end{tabular}
    \caption{Examples of aligned punctuation marks with varying cosine similarities. The aligned tokens are underlined.}
    \label{tab:punc_examples}
\end{table}

\begin{figure}
    \centering
    \includegraphics[width=0.95\linewidth]{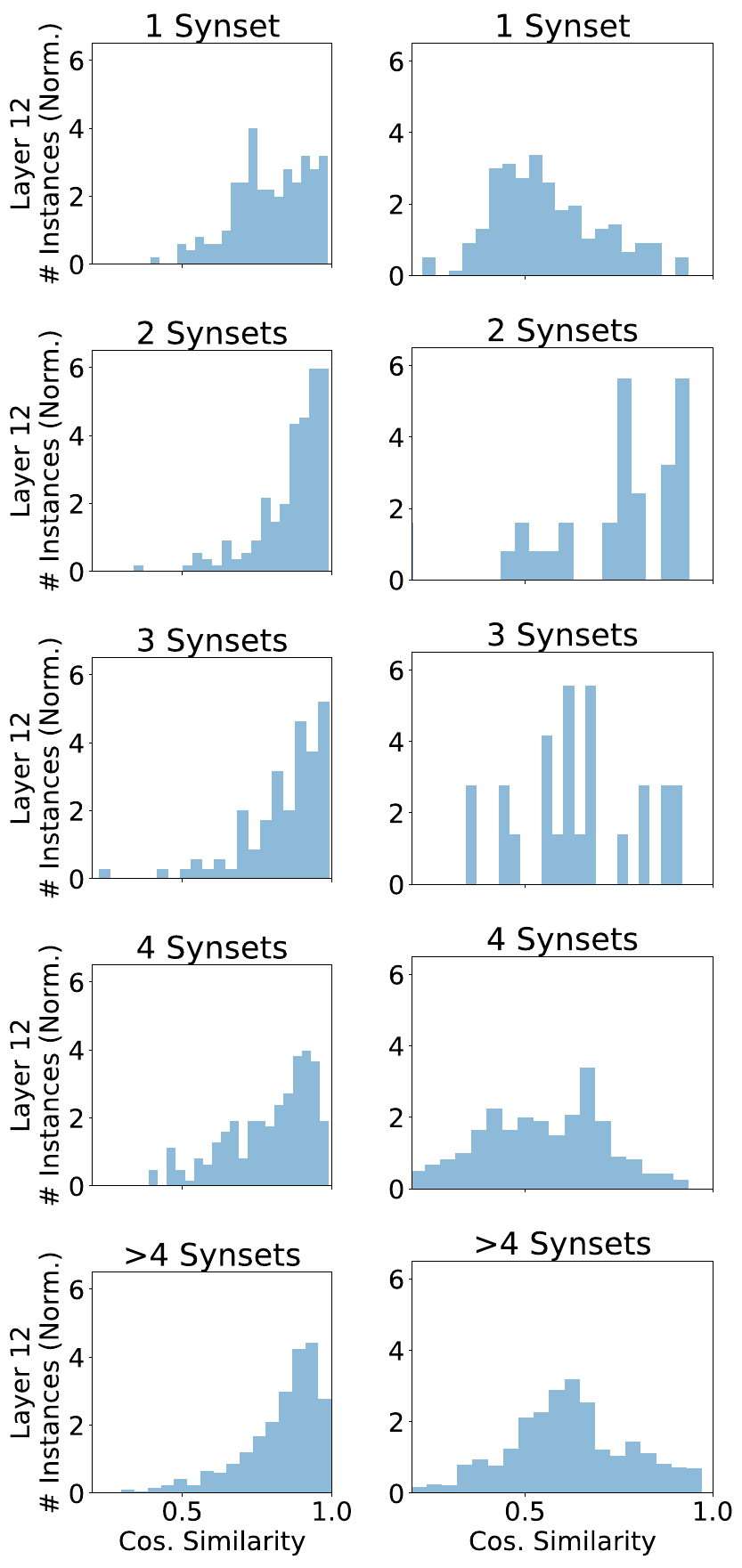}
    \caption{Distributions of cosine similarity for aligned (left) and unaligned (right) same words from the last layer of BERT grouped by the number of senses each word has.
    }
    \label{fig:polysemy}
    \vskip -0.2in
\end{figure}

\subsubsection{Polysemy} \label{sec:polysemy}

Previous work has shown that BERT embeddings form clusters based on word senses \cite{wiedemann2019does}. In the context of aligned words in a paraphrase, we would expect even a highly polysemous word to have similar embeddings in the two phrases.

To measure polysemy, we consider the number of WordNet synsets of a word, focusing on same aligned and same unaligned words. In order to have enough data to make a good comparison, we use the 4,000 sampled paraphrases from Section~\ref{sec:word-level}, as well as an additional random sample of long paraphrases with at least one unaligned same word. We then downsample the aligned same words to get 1,597 instances of both unaligned and aligned same words that are present in WordNet, with up to 52 synsets.\footnote{We look up WordNet synsets using the Python NLTK library \cite{bird2009natural}.}

In Figure~\ref{fig:polysemy}, we show the cosine similarity distributions for both aligned and unaligned words with different levels of polysemy across the last layer of BERT. There is not a substantial difference between words with different synsets, which supports our conclusion that BERT successfully captures the semantics of aligned same words in paraphrases. We do see a difference between aligned and unaligned words.
Aligned words peak at a high cosine similarity, while unaligned words roughly follow a normal distribution centered around 0.5. 
Note that for unaligned words with two or three synsets, there is not enough data to draw conclusions about the cosine similarity distributions. Overall, these plots show that even highly polysemous aligned same words have very similar embeddings in the context of a paraphrase.

\emph{Conclusion: How polysemous a word is does not substantially impact BERT's ability to consistently represent it.}

\subsection{Contextualization in BERT Layers} \label{sec:contextualization}

In this section, we consider how context-specific the embeddings in a paraphrase are.
\citet{ethayarajh-2019-contextual} showed that BERT word embeddings are more context-specific in higher layers.
They measure this using the self-similarity of words, defined as the average cosine similarity between a word's contextualized representations across its unique contexts, and show that self-similarity consistently decreases with higher layers of BERT, indicating that the contextualization of words is increasing.

\begin{figure}
    \centering
    \includegraphics[width=0.95\linewidth]{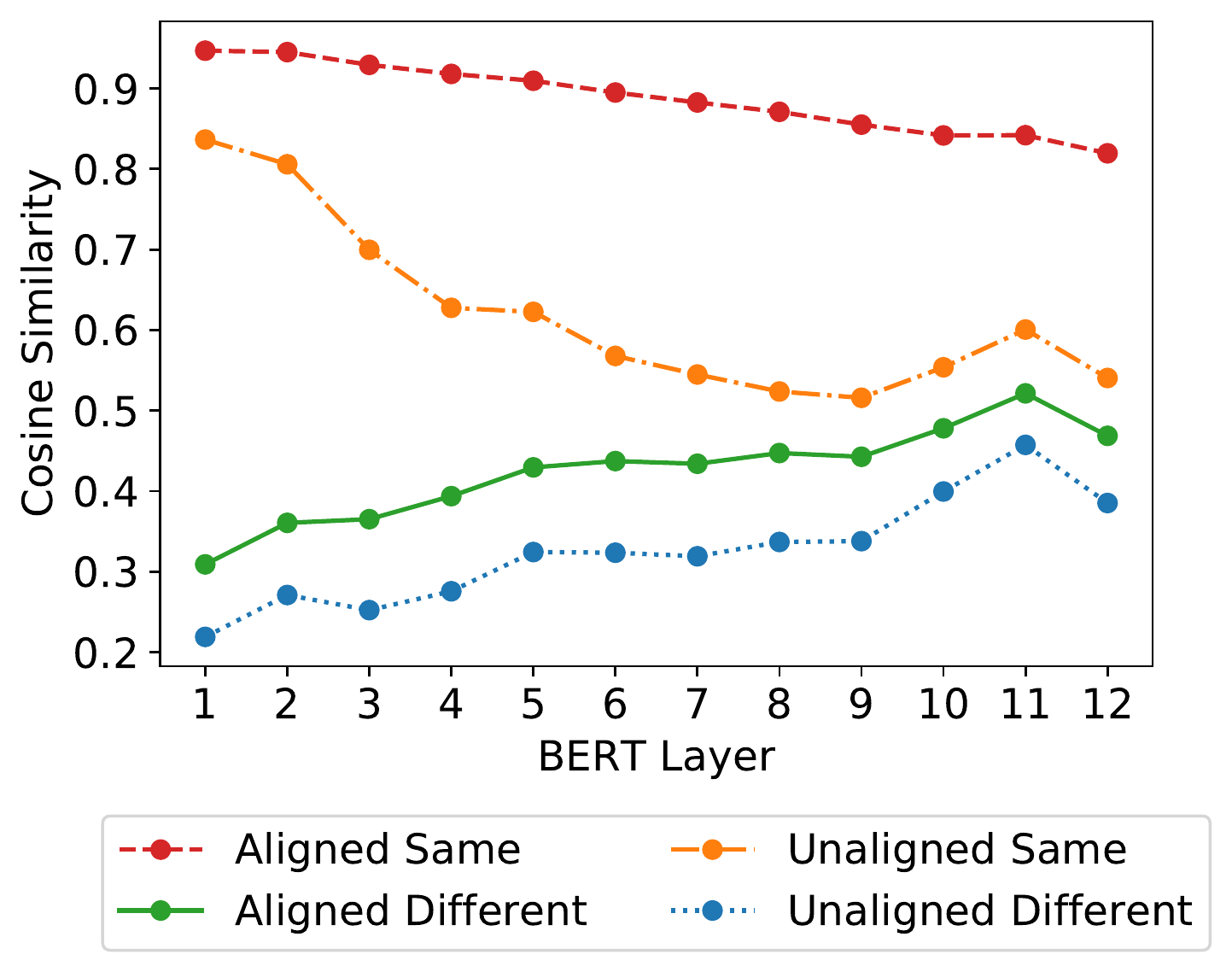}
    \caption{Cosine similarity for different groups of words in the PPDB across all layers of BERT. Decreasing cosine similarity indicates increasing contextualization, and vice versa.}
    \label{fig:contextualized_paraphrases}
\end{figure}

We compare this observation to the paraphrase setting that we have been exploring in this paper. Because there are only two phrases in a paraphrase, we cannot implement the full self-similarity metric.
Instead, we measure the cosine similarity between two words aligned in a paraphrase, shown in Figure~\ref{fig:contextualized_paraphrases}. This revised metric measures is similar to self-similarity.

We see two trends in Figure~\ref{fig:contextualized_paraphrases}. The first, decreasing cosine similarity, is seen with same words, whether aligned or unaligned, and is similar to what \citet{ethayarajh-2019-contextual} report with decreasing self-similarity scores. This trend is stronger with unaligned words than with aligned words, indicating the model is capturing the fact that while these words have the same form, they are being used differently.
The second trend that we see is the opposite, increasing cosine similarity, and we see this trend with different words, both aligned and unaligned. This indicates decreasing contextualization.

\emph{Conclusion: As seen in prior work, the standard way of using vectors from BERT's layers does not capture the same level of contextualization in all layers. However, in contrast to prior work, when controlling for semantics of the context, it seems that later layers are capturing more of the context, appropriately making words less similar when they are being used in different ways.}

\section{Conclusion}

Paraphrases with word alignments are a useful tool for studying the behavior of contextual language models.
In this paper, we used them to study several contextual models, with a particular focus on BERT.
Where possible, we compared our results with prior work, finding patterns that are consistent with the literature.
Specifically, our results confirm that BERT consistently represents paraphrases, even for cases with polysemous words, but that individual word representations are overly sensitive to position, particularly for punctuation.
One exception is that we found that words in a sentence are more similar to each other in later layers of BERT, in contrast to prior work that did not control for meaning using paraphrases. 

The analysis method we introduced opens up new opportunities, such as the comparison of aligned and unaligned, same and different words, which shows the sensitivity of these models to the specific word used.
Paraphrases have the potential to inform exploration of other representation methods, showing which way of using the output of language models most accurately captures semantics consistently.
We hope our findings will inform future work on contextualized models, and the applications that rely on them. 

\bibliography{paper}

\begin{thebibliography}{30}
\expandafter\ifx\csname natexlab\endcsname\relax\def\natexlab#1{#1}\fi

\bibitem[{Bannard and Callison-Burch(2005)}]{10.3115/1219840.1219914}
Colin Bannard and Chris Callison-Burch. 2005.
\newblock \href {https://doi.org/10.3115/1219840.1219914} {Paraphrasing with
  bilingual parallel corpora}.
\newblock In \emph{Proceedings of the 43rd Annual Meeting of the Association
  for Computational Linguistics}, pages 597--604.

\bibitem[{Bird et~al.(2009)Bird, Klein, and Loper}]{bird2009natural}
Steven Bird, Ewan Klein, and Edward Loper. 2009.
\newblock \emph{Natural language processing with Python: {A}nalyzing text with
  the natural language toolkit}.
\newblock O'Reilly Media, Inc.

\bibitem[{Chan and Fan(2019)}]{chan-fan-2019-bert}
Ying-Hong Chan and Yao-Chung Fan. 2019.
\newblock \href {https://doi.org/10.18653/v1/W19-8624} {{BERT} for question
  generation}.
\newblock In \emph{Proceedings of the 12th International Conference on Natural
  Language Generation}, pages 173--177.

\bibitem[{Devlin et~al.(2019)Devlin, Chang, Lee, and
  Toutanova}]{devlin2018bert}
Jacob Devlin, Ming-Wei Chang, Kenton Lee, and Kristina Toutanova. 2019.
\newblock \href {https://aclanthology.org/N19-1423/} {{BERT}: Pre-training of
  deep bidirectional transformers for language understanding}.
\newblock In \emph{Proceedings of the 2019 Conference of the North American
  Chapter of the Association for Computational Linguistics: Human Language
  Technologies}, pages 4171--4186.

\bibitem[{Ethayarajh(2019)}]{ethayarajh-2019-contextual}
Kawin Ethayarajh. 2019.
\newblock \href {https://doi.org/10.18653/v1/D19-1006} {How contextual are
  contextualized word representations? {C}omparing the geometry of {BERT},
  {ELM}o, and {GPT}-2 embeddings}.
\newblock In \emph{Proceedings of the 2019 Conference on Empirical Methods in
  Natural Language Processing and the 9th International Joint Conference on
  Natural Language Processing}, pages 55--65.

\bibitem[{Fellbaum(1998)}]{fellbaum1998wordnet}
Christiane Fellbaum. 1998.
\newblock \href
  {https://onlinelibrary.wiley.com/doi/full/10.1002/9781405198431.wbeal1285}
  {\emph{WordNet}}.
\newblock Wiley Online Library.

\bibitem[{Ganitkevitch et~al.(2013)Ganitkevitch, Van~Durme, and
  Callison-Burch}]{ganitkevitch-etal-2013-ppdb}
Juri Ganitkevitch, Benjamin Van~Durme, and Chris Callison-Burch. 2013.
\newblock \href {https://www.aclweb.org/anthology/N13-1092} {{PPDB}: The
  paraphrase database}.
\newblock In \emph{Proceedings of the 2013 Conference of the North {A}merican
  Chapter of the Association for Computational Linguistics: Human Language
  Technologies}, pages 758--764.

\bibitem[{Gar{\'\i}~Soler and
  Apidianaki(2020)}]{gari-soler-apidianaki-2020-bert}
Aina Gar{\'\i}~Soler and Marianna Apidianaki. 2020.
\newblock \href {https://doi.org/10.18653/v1/2020.emnlp-main.598} {{BERT} knows
  punta cana is not just beautiful, it{'}s gorgeous: Ranking scalar adjectives
  with contextualised representations}.
\newblock In \emph{Proceedings of the 2020 Conference on Empirical Methods in
  Natural Language Processing (EMNLP)}, pages 7371--7385.

\bibitem[{Huang et~al.(2019)Huang, Sun, Qiu, and
  Huang}]{huang-etal-2019-glossbert}
Luyao Huang, Chi Sun, Xipeng Qiu, and Xuanjing Huang. 2019.
\newblock \href {https://doi.org/10.18653/v1/D19-1355} {{G}loss{BERT}: {BERT}
  for word sense disambiguation with gloss knowledge}.
\newblock In \emph{Proceedings of the 2019 Conference on Empirical Methods in
  Natural Language Processing and the 9th International Joint Conference on
  Natural Language Processing (EMNLP-IJCNLP)}, pages 3509--3514.

\bibitem[{Kovaleva et~al.(2019)Kovaleva, Romanov, Rogers, and
  Rumshisky}]{kovaleva-etal-2019-revealing}
Olga Kovaleva, Alexey Romanov, Anna Rogers, and Anna Rumshisky. 2019.
\newblock \href {https://doi.org/10.18653/v1/D19-1445} {Revealing the dark
  secrets of {BERT}}.
\newblock In \emph{Proceedings of the 2019 Conference on Empirical Methods in
  Natural Language Processing and the 9th International Joint Conference on
  Natural Language Processing (EMNLP-IJCNLP)}, pages 4365--4374.

\bibitem[{Mahmoud and Torki(2020)}]{mahmoud-torki-2020-alexu}
Somaia Mahmoud and Marwan Torki. 2020.
\newblock \href {https://www.aclweb.org/anthology/2020.semeval-1.33}
  {{A}lex{U}-{AUX}-{BERT} at {S}em{E}val-2020 task 3: Improving {BERT}
  contextual similarity using multiple auxiliary contexts}.
\newblock In \emph{Proceedings of the Fourteenth Workshop on Semantic
  Evaluation}, pages 270--274. International Committee for Computational
  Linguistics.

\bibitem[{Mickus et~al.(2020)Mickus, Paperno, Constant, and van
  Deemter}]{mickus2019you}
Timothee Mickus, Denis Paperno, Mathieu Constant, and Kees van Deemter. 2020.
\newblock \href {https://aclanthology.org/2020.scil-1.35} {What do you mean,
  {BERT}?}
\newblock In \emph{Proceedings of the Society for Computation in Linguistics
  2020}, pages 279--290.

\bibitem[{Mikolov et~al.(2013)Mikolov, Le, and
  Sutskever}]{mikolov2013exploiting}
Tomas Mikolov, Quoc~V Le, and Ilya Sutskever. 2013.
\newblock \href {https://arxiv.org/abs/1309.4168} {Exploiting similarities
  among languages for machine translation}.
\newblock \emph{arXiv preprint arXiv:1309.4168}.

\bibitem[{Parker et~al.(2011)Parker, Graff, Kong, Chen, and
  Maeda}]{graff2003english}
Robert Parker, David Graff, Junbo Kong, Ke~Chen, and Kazuaki Maeda. 2011.
\newblock \href {https://catalog.ldc.upenn.edu/LDC2011T07} {English gigaword
  fifth edition}.
\newblock \emph{Linguistic Data Consortium, Philadelphia}.

\bibitem[{Pavlick et~al.(2015)Pavlick, Rastogi, Ganitkevitch, Van~Durme, and
  Callison-Burch}]{pavlick-etal-2015-ppdb}
Ellie Pavlick, Pushpendre Rastogi, Juri Ganitkevitch, Benjamin Van~Durme, and
  Chris Callison-Burch. 2015.
\newblock \href {https://doi.org/10.3115/v1/P15-2070} {{PPDB} 2.0: Better
  paraphrase ranking, fine-grained entailment relations, word embeddings, and
  style classification}.
\newblock In \emph{Proceedings of the 53rd Annual Meeting of the Association
  for Computational Linguistics and the 7th International Joint Conference on
  Natural Language Processing}, pages 425--430.

\bibitem[{Rastogi et~al.(2015)Rastogi, Van~Durme, and
  Arora}]{rastogi2015multiview}
Pushpendre Rastogi, Benjamin Van~Durme, and Raman Arora. 2015.
\newblock \href {https://doi.org/10.3115/v1/N15-1058} {Multiview {LSA}:
  Representation learning via generalized {CCA}}.
\newblock In \emph{Proceedings of the 2015 Conference of the North {A}merican
  Chapter of the Association for Computational Linguistics: Human Language
  Technologies}, pages 556--566.

\bibitem[{Reimers and Gurevych(2019)}]{reimers-gurevych-2019-sentence}
Nils Reimers and Iryna Gurevych. 2019.
\newblock \href {https://doi.org/10.18653/v1/D19-1410} {Sentence-{BERT}:
  Sentence embeddings using {S}iamese {BERT}-networks}.
\newblock In \emph{Proceedings of the 2019 Conference on Empirical Methods in
  Natural Language Processing and the 9th International Joint Conference on
  Natural Language Processing (EMNLP-IJCNLP)}, pages 3982--3992.

\bibitem[{Rogers et~al.(2020)Rogers, Kovaleva, and
  Rumshisky}]{rogers2020primer}
Anna Rogers, Olga Kovaleva, and Anna Rumshisky. 2020.
\newblock \href {https://doi.org/10.1162/tacl_a_00349} {A primer in
  {BERT}ology: What we know about how {BERT} works}.
\newblock \emph{Transactions of the Association for Computational Linguistics},
  8:842--866.

\bibitem[{Rush et~al.(2015)Rush, Chopra, and Weston}]{Rush_2015}
Alexander~M. Rush, Sumit Chopra, and Jason Weston. 2015.
\newblock \href {https://doi.org/10.18653/v1/d15-1044} {A neural attention
  model for abstractive sentence summarization}.
\newblock \emph{Proceedings of the 2015 Conference on Empirical Methods in
  Natural Language Processing}.

\bibitem[{Sellam et~al.(2020)Sellam, Das, and Parikh}]{sellam-etal-2020-bleurt}
Thibault Sellam, Dipanjan Das, and Ankur Parikh. 2020.
\newblock \href {https://doi.org/10.18653/v1/2020.acl-main.704} {{BLEURT}:
  Learning robust metrics for text generation}.
\newblock In \emph{Proceedings of the 58th Annual Meeting of the Association
  for Computational Linguistics}, pages 7881--7892.

\bibitem[{Soler-Company and Wanner(2017)}]{soler-company-wanner-2017-relevance}
Juan Soler-Company and Leo Wanner. 2017.
\newblock \href {https://www.aclweb.org/anthology/E17-2108} {On the relevance
  of syntactic and discourse features for author profiling and identification}.
\newblock In \emph{Proceedings of the 15th Conference of the {E}uropean Chapter
  of the Association for Computational Linguistics}, pages 681--687.

\bibitem[{Spearman(1910)}]{spearman1910correlation}
Charles Spearman. 1910.
\newblock Correlation calculated from faulty data.
\newblock \emph{British Journal of Psychology, 1904-1920}, 3(3):271--295.

\bibitem[{Tatman and Paullada(2017)}]{tatman-punctuation}
Rachael Tatman and Amandalynne Paullada. 2017.
\newblock Social identity and punctuation variation in the {\#bluelivesmatter}
  and {\#blacklivesmatter} twitter communities.
\newblock In \emph{The 33rd Northwest Linguistics Conference}.

\bibitem[{Tiedemann(2012)}]{tiedemann-2012-parallel}
J{\"o}rg Tiedemann. 2012.
\newblock \href
  {http://www.lrec-conf.org/proceedings/lrec2012/pdf/463_Paper.pdf} {Parallel
  data, tools and interfaces in {OPUS}}.
\newblock In \emph{Proceedings of the Eighth International Conference on
  Language Resources and Evaluation ({LREC}'12)}, pages 2214--2218. European
  Language Resources Association (ELRA).

\bibitem[{Tsvetkov et~al.(2016)Tsvetkov, Faruqui, Ling, MacWhinney, and
  Dyer}]{tsvetkov-etal-2016-learning}
Yulia Tsvetkov, Manaal Faruqui, Wang Ling, Brian MacWhinney, and Chris Dyer.
  2016.
\newblock \href {https://doi.org/10.18653/v1/P16-1013} {Learning the curriculum
  with {B}ayesian optimization for task-specific word representation learning}.
\newblock In \emph{Proceedings of the 54th Annual Meeting of the Association
  for Computational Linguistics}, pages 130--139.

\bibitem[{Wiedemann et~al.(2019)Wiedemann, Remus, Chawla, and
  Biemann}]{wiedemann2019does}
Gregor Wiedemann, Steffen Remus, Avi Chawla, and Chris Biemann. 2019.
\newblock \href
  {https://www.inf.uni-hamburg.de/en/inst/ab/lt/publications/2019-wiedemannetal-konvens-bert.pdf}
  {Does {BERT} make any sense? {I}nterpretable word sense disambiguation with
  contextualized embeddings}.
\newblock In \emph{Konferenz zur Verarbeitung {natürlicher} Sprache /
  Conference on Natural Language Processing}.

\bibitem[{{Yi} and {Tao}(2019)}]{8682260}
J.~{Yi} and J.~{Tao}. 2019.
\newblock \href {https://ieeexplore.ieee.org/document/8682260} {Self-attention
  based model for punctuation prediction using word and speech embeddings}.
\newblock In \emph{IEEE International Conference on Acoustics, Speech and
  Signal Processing}, pages 7270--7274.

\bibitem[{Yoosuf and Yang(2019)}]{yoosuf-yang-2019-fine}
Shehel Yoosuf and Yin Yang. 2019.
\newblock \href {https://doi.org/10.18653/v1/D19-5011} {Fine-grained propaganda
  detection with fine-tuned {BERT}}.
\newblock In \emph{Proceedings of the Second Workshop on Natural Language
  Processing for Internet Freedom: Censorship, Disinformation, and Propaganda},
  pages 87--91.

\bibitem[{Yu and Ettinger(2020)}]{compositionality-test}
Lang Yu and Allyson Ettinger. 2020.
\newblock \href {https://doi.org/10.18653/v1/2020.emnlp-main.397} {Assessing
  phrasal representation and composition in transformers}.
\newblock In \emph{Proceedings of the 2020 Conference on Empirical Methods in
  Natural Language Processing (EMNLP)}, pages 4896--4907.

\bibitem[{Zhang* et~al.(2020)Zhang*, Kishore*, Wu*, Weinberger, and
  Artzi}]{zhang2019bertscore}
Tianyi Zhang*, Varsha Kishore*, Felix Wu*, Kilian~Q. Weinberger, and Yoav
  Artzi. 2020.
\newblock \href {https://openreview.net/forum?id=SkeHuCVFDr} {Bertscore:
  Evaluating text generation with bert}.
\newblock In \emph{International Conference on Learning Representations}.

\end{thebibliography}
\bibliographystyle{acl_natbib}

\newpage

\appendix

\section{Extra breakdowns of results}

Table~\ref{tab:length-by-layer} presents an expanded version of Table~\ref{tab:length_breakdown}, with results for each layer of BERT.
All layers perform better with longer paraphrases, but the improvement is largest for the last layer

\begin{figure}[h!]
    \centering
    \includegraphics[width=0.95\linewidth]{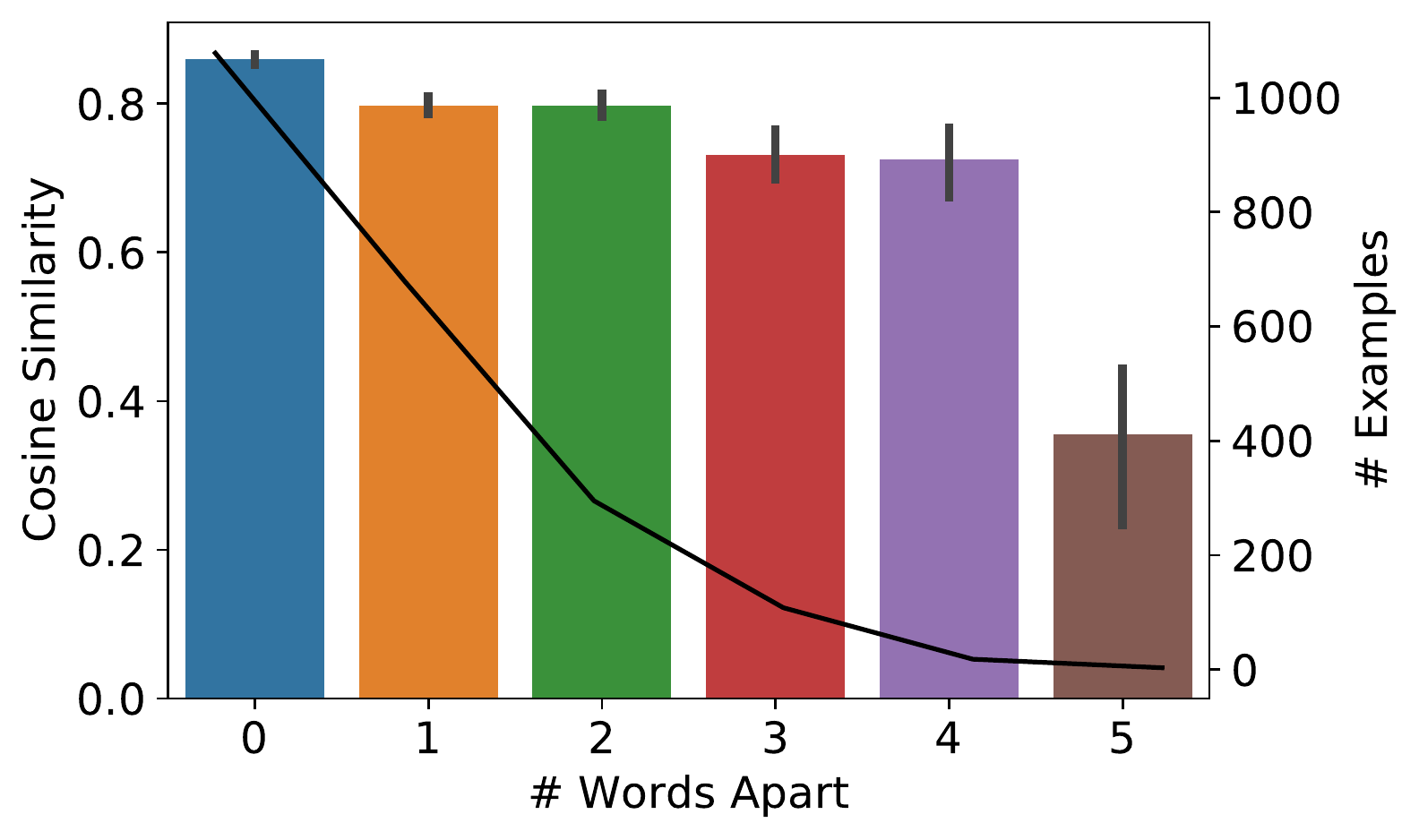}
    \caption{Cosine similarity using the last layer of BERT for aligned same words broken down by the number of words apart the words are in the two phrases (shown in bar plot and left y-axis). Error bars indicate confidence intervals. The line graph and the right y-axis show how many examples we have for each category.}
    \label{fig:words_apart}
\end{figure}

Figure~\ref{fig:words_apart} shows the specific values for similarity broken down by distance apart of words in the phrases.
This shows the pattern of decreasing similarity as words are further away.

\begin{table}[h!]
    \small
    \centering
    \setlength{\tabcolsep}{5pt}
    \begin{tabular}{lrrrr}
        \toprule
        & \multicolumn{3}{c}{Average Length} & All \\
        Method & 1-2.5 & 2.5-4 & 4-6 & \\
        \midrule
        BERT Layer 1 & $0.18$ & $0.35$ & $0.47$ & $0.34$ \\
        BERT Layer 2 & $0.18$ & $0.35$ & $0.49$ & $0.33$ \\
        BERT Layer 3 & $0.18$ & $0.37$ & $0.48$ & $0.31$ \\
        BERT Layer 4 & $0.18$ & $0.38$ & $0.48$ & $0.3$ \\
        BERT Layer 5 & $0.18$ & $0.39$ & $0.48$ & $0.3$ \\
        BERT Layer 6 & $0.19$ & $0.39$ & $0.49$ & $0.29$ \\
        BERT Layer 7 & $0.2$ & $0.4$ & $0.49$ & $0.3$ \\
        BERT Layer 8 & $0.21$ & $0.4$ & $0.5$ & $0.28$ \\
        BERT Layer 9 & $0.21$ & $0.4$ & $0.5$ & $0.28$ \\
        BERT Layer 10 & $0.22$ & $0.38$ & $0.48$ & $0.29$ \\
        BERT Layer 11 & $0.22$ & $0.36$ & $0.46$ & $0.29$ \\
        BERT Layer 12 & $0.1$ & $0.35$ & $0.51$ & $0.16$ \\
        BERT Concat. & $0.2$ & $0.4$ & $0.51$ & $0.31$ \\
        \midrule
        w2v Average & $0.35$ & $0.32$ & $0.41$ & $0.43$\\
        \midrule
        PPDB model & $0.41$ & $0.50$ & $0.51$ &$0.50$ \\
        \midrule
        \midrule
        Num. phrases & $17,517$ & $5,349$ & $2,870$ & $25,736$\\
        Avg. human & $2.40$ & $2.94$ & $3.26$ & $2.60$ \\
        \bottomrule
    \end{tabular}
    \caption{\label{tab:length-by-layer}This is a version of Table~\ref{tab:length_breakdown} with per-layer results. Spearman's $\rho$ between human-annotated PPDB paraphrases and different embedding methods (BERT and w2v), broken down by average paraphrase length (the average number of words in each of the two phrases in the paraphrase). At the bottom of the table, we include the length distribution of the human-annotated paraphrases, as well as the average human annotation for each set of grouped lengths.}
\end{table}

\end{document}